\documentclass[letterpaper]{article} % DO NOT CHANGE THIS
\usepackage{aaai24}  % DO NOT CHANGE THIS
\usepackage{times}  % DO NOT CHANGE THIS
\usepackage{helvet}  % DO NOT CHANGE THIS
\usepackage{courier}  % DO NOT CHANGE THIS
\usepackage[hyphens]{url}  % DO NOT CHANGE THIS
\usepackage{graphicx} % DO NOT CHANGE THIS
\usepackage{natbib}  % DO NOT CHANGE THIS AND DO NOT ADD ANY OPTIONS TO IT
\usepackage{caption} % DO NOT CHANGE THIS AND DO NOT ADD ANY OPTIONS TO IT
\usepackage{algorithm}
\usepackage{algpseudocode}
\usepackage{amssymb}
\usepackage[centertags]{amsmath}

\usepackage{newfloat}
\usepackage{listings}
\DeclareCaptionStyle{ruled}{labelfont=normalfont,labelsep=colon,strut=off} % DO NOT CHANGE THIS
\floatstyle{ruled}
\newfloat{listing}{tb}{lst}{}
\floatname{listing}{Listing}
\title{Personalized Interpretation on Federated Learning: A Virtual Concepts approach}
\author{
    Peng Yan, Guodong Long, Jing Jiang, Michael Blumenstein
}
\affiliations{
    University of Technology Sydney
}

\usepackage{bibentry}
\begin{document}

\maketitle

\begin{abstract}
Tackling non-IID data is an open challenge in federated learning research. Existing FL methods, including robust FL and personalized FL, are designed to improve model performance without consideration of interpreting non-IID across clients. This paper aims to design a novel FL method to robust and interpret the non-IID data across clients. Specifically, we interpret each client's dataset as a mixture of conceptual vectors that each one represents an interpretable concept to end-users. These conceptual vectors could be pre-defined or refined in a human-in-the-loop process or be learnt via the optimization procedure of the federated learning system. In addition to the interpretability, the clarity of client-specific personalization could also be applied to enhance the robustness of the training process on FL system. The effectiveness of the proposed method have been validated on benchmark datasets.
\end{abstract}

\section{Motivation}
A critical challenge in PerFL is the absence of well-defined concepts of personalisation. Client preferences and personalised properties are implied in training data and enclosed on each client. They could be a client's favour towards specific classes or a specific noise mixed up with input features. The only tangible information is the shift in data distribution across clients.

Meanwhile, most machine learning models, e.g., DNNs, are trained in an end-to-end paradigm. They are optimised by back-propagating supervised information, e.g., classification loss, from the output layer to the input layer. Personalisation is performed indirectly when a model is tuned for tasks like classification. This learning schema is less efficient in PerFL. The on-device training tends to overfit a client's local data due to limited and unbalanced training samples. The aggregation step on the server, in turn, will neutralise personalised information when synthesising the global model, e.g., by averaging local updates.

However, it is worth noting that though there is no supervised information of significantly defined client properties, a feature distinguishing model personalisation from unsupervised tasks is that data in PerFL are explicitly partitioned. Samples from the same client will demonstrate a client-specific bias toward certain properties. Then, one may assume that there were invisible labels of clients inducing the on-device training to progress toward a client's preferences, i.e., personalisation. The client-based data partition essentially supervises PerFL's training process, so this research calls the learning paradigm Client-Supervised Learning.

Based on the thought above, this research introduces Virtual Concepts (VC) to explicate client-supervised information. The VCs are representations of potential structure information extracted from training data. They can be learned independently of downstream classification tasks by a novel FedVC algorithm, which facilitates understanding client properties and boosts model personalisation.

\begin{figure}[h]
    \centering
    \includegraphics[width=0.9\columnwidth]{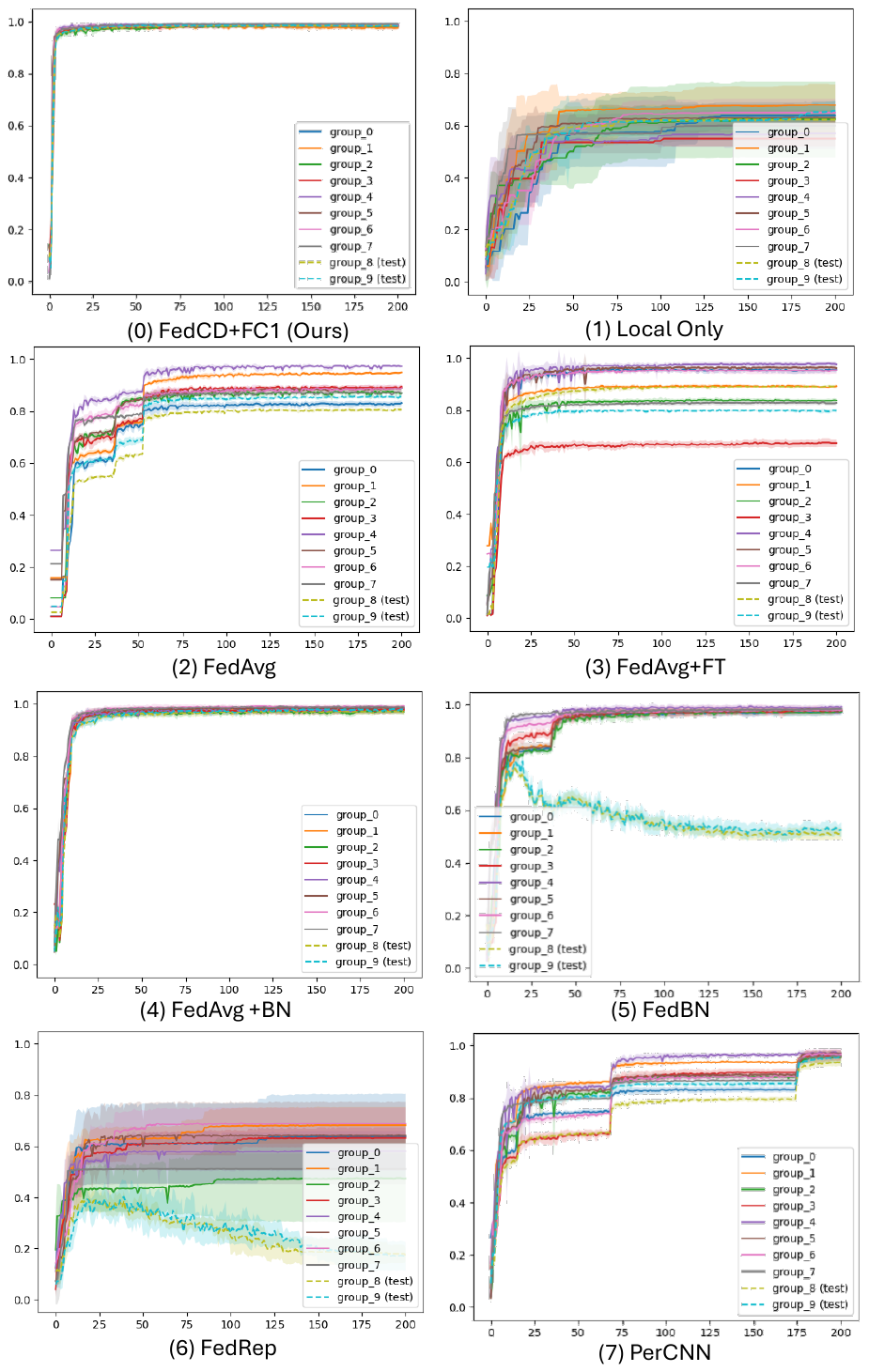}
    \caption{Illustration to FedVC. (a) data distribution in an FL system; (b) virtual concepts (pentagon, plus and triangle) are vectors indicating underlying cluster structures of data, e.g., cluster centres; (c) a client's preference (star) is represented by a combination of virtual concepts; (d) client-supervised loss requires sample representations on the same client (data points within the circle) to be close to each other as they share the identical client preference.}
    \label{chpt6:fig1: FedVC illustration}
\end{figure}

Specifically, FedVC assumes that there is a set of vectors (virtual concepts), each describing a type of client property. A client's preferences are then represented by a combination of VCs, which will be utilised as supervised information to guide the training progress of the global model. \textbf{Figure~\ref{chpt6:fig1: FedVC illustration}} gives an illustration to the propose FedVC.
To learn the VCs, FedVC evaluates the underlying distribution structure in data by formulating the learning task into a Gaussian Mixture Model (GMM) that can be solved by most unsupervised learning methods, e.g., Expectation-Maximisation algorithm (EM).

Experiments on real-world datasets show that the VCs can work as supervised information to train a robust global model to the changing distributions. Further study demonstrates that the VCs are useful in exploring meaningful client properties by discovering distribution structures implied in training data.

The main contributions are summarised as follows:
\begin{itemize}
\item The research proposes virtual concepts describing client preferences. The VCs are representations of distribution structure extracted from training data. They provide us with a way to explore meaningful client properties relevant to model personalisation.

\item The research proposes a novel client-supervised PerFL framework that utilises virtual concept vectors as supervised information to train the global model. The VCs will allow an FL algorithm to simultaneously learn class and client knowledge so that the learned global model can achieve on-deployment personalisation, where the global model will not require an extra fine-tuning process at the test stage.

\item The research formulates the learning task of VCs into a Gaussian Mixture Model that most unsupervised learning methods can solve. The proposed FedVC framework is compatible with most FL methods, where they can be integrated as an add-on to improve personalisation performance and model interpretability.

\item Contrast with baseline methods shows that FL models trained with VCs can simultaneously learn class and client knowledge. It achieves competitive personalisation performance without requiring extra fine-tuning steps or personal parameters.

\item Empirical studies show that VCs can discover meaningful distribution structures implied in training, facilitating the uncovering of client properties related to model personalisation.
\end{itemize}

\section{Methodology}
\subsection{Client-supervised PerFL}
\label{Client-supervised PerFL}
Let $\mathcal{C}=\{c_1,...,c_M\}$ denote $m$ virtual concept vectors, a client's preference is then represented by $p^{(k)}=\sum_{m=1}^{M}\upsilon^{(k)}_{m}c_{m}$, where $k$ is the client index, and $\upsilon_m$ is a factor measuring the degree the client relevant to $c_m$, i.e., how typical the client has the property of $c_m$. FedVC aims to utilise $p^{(k)}$ as supervised information to guide FL's learning process so that the global model can learn client knowledge explicitly. 

\begin{figure}[h]
    \centering
    \includegraphics[width=0.4\columnwidth]{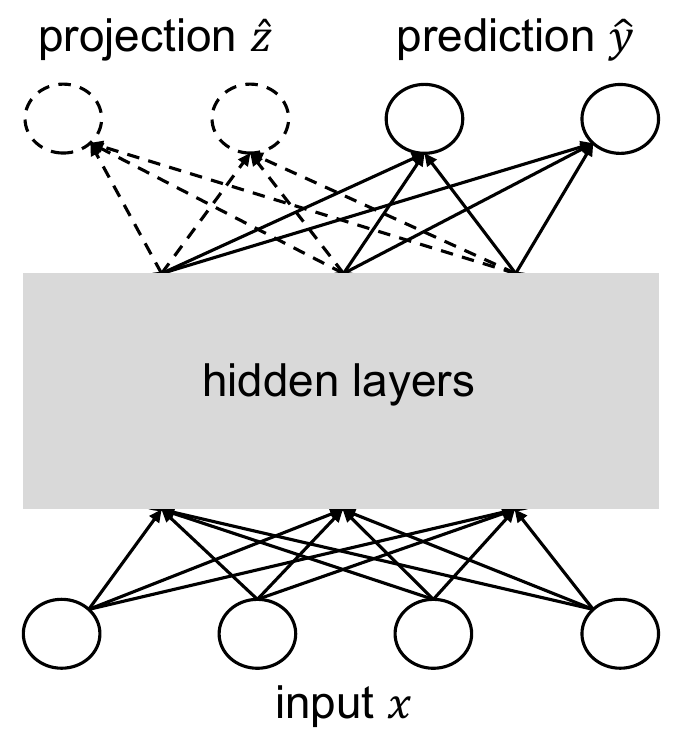}
    \caption{Projection head}
    \label{chpt6:fig2: projection head}
\end{figure}
Specifically, FedVC adds a projection head to FL's global model to extract a representation $\hat{z}_{i}^{(k)}$ of potential client properties (see \textbf{Figure~\ref{chpt6:fig2: projection head}}). One can evaluate a sample's relevance to each concept by a similarity function, e.g., \textbf{Equation~\ref{eq: sample relevance to vc}}, and derive an estimated client preference $\hat{p}_{i}^{(k)}=\sum_{m=1}^{M}\hat{s}_{i,m}^{(k)}c_{m}$, 
where $i$ is the sample index and $\iota$ is a hyperparameter.
\begin{equation}
\label{eq: sample relevance to vc}
s_{i,m}^{(k)}=\frac{\upsilon_{m}^{(k)}\text{exp}(-\iota\|\hat{z}_i^{(k)}-c_m\|^{2})}{\sum_{m=1}^{M}\upsilon_{m}^{(k)}\text{exp}(-\iota\|\hat{z}_i^{(k)}-c_m\|^{2})}    
\end{equation}
Then, there will be a supervised loss regarding client preferences, i.e., $l_p(\hat{p}^{(k)},p_i^{(k)})=\|\hat{p}^{(k)}-p_{i}^{(k)}\|^{2}$. It can be integrated into any FL framework and solved by gradient-based methods. Details of the learning algorithm are in \textbf{Algorithm~\ref{chapter6:alg1: FedVC}}.

\subsection{Virtual Concepts}
As virtual concepts correspond to client properties, a sample is then assumed to be generated by some random process involving a mixture of multiple client properties. FedVC formulates the assumption into a Gaussian Mixture Model (GMM). For any sample $\mathbf{z}^{(k)}$ on the $k$-th client, there is 
\begin{equation}
    \mathbf{z}^{(k)}\sim \mathcal{P}^{(k)}(\mathbf{z})=\sum_{m=1}^{M}\upsilon^{(k)}_m\mathcal{N}(\mathbf{z};c_m,\Sigma_m)
\end{equation}
where the covariance $\Sigma_m$ is set to be the identity matrix $I$ for simplicity.

Let $\mathcal{C}=\{c_1,...,c_M\}$ denotes the set of VCs and $\Upsilon=\{\{\upsilon^{(1)}_m\}_{m=1}^{M},...,\{\upsilon^{(K)}_m\}_{m=1}^{M}\}$ denotes the set of client preferences, the collaborative learning task for $\mathcal{C}$ and $\Upsilon$ is formulated as 
\begin{equation}
    \label{eq: EM loss}
    \mathcal{C}^{*}, \Upsilon^{*} =\arg\max_{\mathcal{C}, \Upsilon}\sum_{k=1}^{K}\sum_{i=1}^{N_k}\log\mathcal{P}^{(k)}(z_i^{(k)})
\end{equation}
FedVC solves it by the EM framework bellow:
\begin{itemize}
    \item \textbf{E-step}: Given $\mathcal{C}$ and $\Upsilon$, clients estimate local samples' $s_{i,m}^{(k)}$ by \textbf{Equation}~\ref{eq: sample relevance to vc}
    \item \textbf{M-step}: Clients update $\mathcal{C}$ and $\Upsilon$ collaboratively by \textbf{Equation}~\ref{eq: update upsilon} and \textbf{Equation}~\ref{eq: update c_m}
\begin{equation}
    \label{eq: update upsilon}
    \upsilon_{m}^{(k)}=\frac{1}{N_k}\sum_{i=1}^{N_k}s_{i,m}^{(k)}
\end{equation}
\begin{equation}
    \label{eq: update c_m}
    c_{m}=\frac{\sum_{k=1}^{K}\sum_{i=1}^{N_k}s_{i,m}^{(k)}\hat{z}_{i}^{(k)}}{\sum_{k=1}^{K}\sum_{i=1}^{N_k}s_{i,m}^{(k)}}
\end{equation}
\end{itemize}
However, \textbf{Equation~\ref{eq: update upsilon}} and \textbf{Equation~\ref{eq: update c_m}} cannot be applied directly when working with minibatches in FL settings. FedVC uses exponential moving averages as an alternative:
\begin{equation}
    \label{eq: S moving average}
    S_{m}^{'(k)} = S_{m}^{(k)} * \kappa + \sum_{i\in\mathbb{B}}s_{i,m}^{(k)} * (1-\kappa)
\end{equation}

\begin{equation}
    \label{eq: C moving average}
    C_m^{'(k)}=C_m^{(k)} * \kappa + \sum_{i\in\mathbb{B}}s_{i,m}^{(k)}\hat{z}_{i}^{(k)} * (1 - \kappa)
\end{equation}

\begin{equation}
    \label{eq: N moving average}
    N'_k = N_k * \kappa + |\mathbb{B}| * (1-\kappa)
\end{equation}
where $\mathbb{B}$ denotes a minibatch of samples, $|\mathbb{B}|$ denotes the batch size, and $\kappa$ is a smoothing hyperparameter between 0 and 1. Then,
\begin{equation}
    \label{eq: update upsilon moving average}
    \upsilon_{m}^{(k)}=\frac{S_{m}^{'(k)}}{N'_k}
\end{equation}
\begin{equation}
    \label{eq: update c_m moving average}
    c_{m}=\frac{\sum_{k=1}^{K}C_m^{'(k)}}{\sum_{k=1}^{K}S_{m}^{'(k)}}
\end{equation}
The overall learning algorithm is described in \textbf{Algorithm~\ref{chapter6:alg1: FedVC}}.

\subsubsection{Unified Learning Process}
It is worth noting that the client preference $p^{(k)}=\sum_{m=1}^{M}\upsilon^{(k)}_{m}c_{m}$ can be viewed a function of virtual concepts $\mathcal{C}$, so does the loss $l_p(\hat{p}, p)$. Then, the learning processes for $\mathcal{C}$ and the global $\omega$ can be formulated into a unified optimisation task that can be solved in an end-to-end manner, rather than in an alternate way as EM-based methods. 

\begin{figure}[h]
    \centering
    \includegraphics[width=0.9\columnwidth]{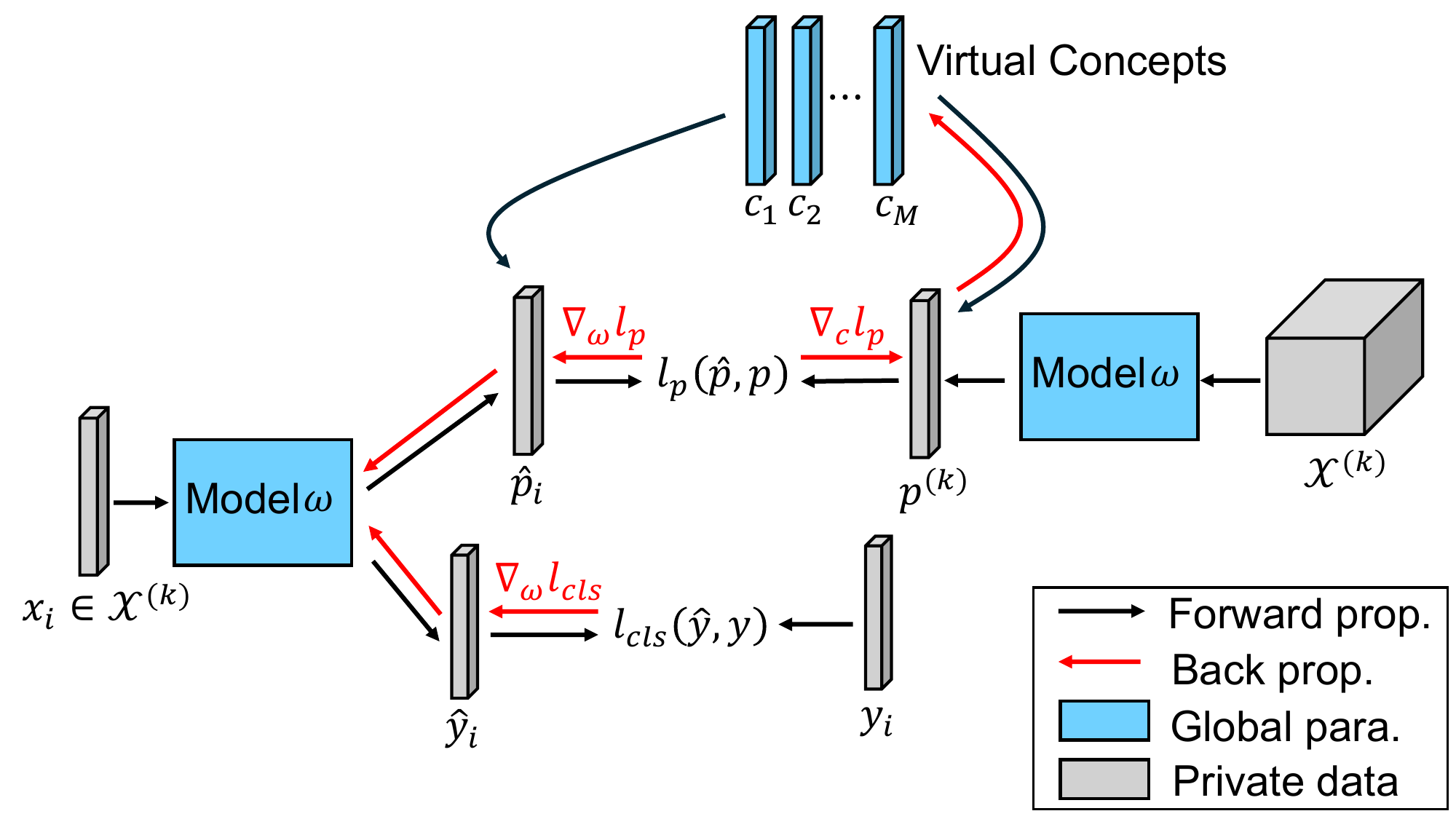}
    \caption{FedVC architecture.}
    \label{chpt6:fig3: FedVC with stopgradient}
\end{figure}
Concretely, as described in \textbf{Figure~\ref{chpt6:fig3: FedVC with stopgradient}}, $l_p(\hat{p}, p)$ will simultaneously provide supervised information for optimising virtual concepts and the model. The unified learning object is formulated as 
\begin{equation}
    \label{eq: fedvc}
    \begin{aligned}
    \omega^{*}, \mathcal{C}^{*} =\arg\min_{\omega, \mathcal{C}}\sum_{k=1}^{K}\alpha_k \mathcal{L}_k(\omega, \mathcal{C})
    \end{aligned}
\end{equation}
where
\begin{equation}
\begin{aligned}
    \mathcal{L}_k(\omega)=&(1/N_k)\sum_{i=1}^{N_k}l_{cls}(\hat{y}_{i}^{(k)}, y_{i}^{(k)}) \\
    &+ l_{p}(\hat{p}_{i}^{(k)}, \text{sg}[p^{(k)}]) + \gamma l_{p}(\text{sg}[\hat{p}_{i}^{(k)}], p^{(k)})
\end{aligned}
\label{eq: fedvc loss}
\end{equation}
The $\text{sg}[\cdot]$ is the stopgradient operator~\cite{van2017neural}, where the operand will feed forward as normal but have zero partial derivatives, being a non-updated constant. $\gamma$ is a hyperparameter balancing the two loss. The corresponding learning process is summarised in \textbf{Algorithm~\ref{chapter6:alg1: FedVC end-to-end}}.

\section{Experiments}
This section empirical studies the advantages of FedVC in learning from clients with non-I.I.D. data. The FedVC can learn a robust FL global model for the changing data distributions of unseen/test clients. The FedVC's global model can be directly deployed to the test clients while achieving comparable performance to other personalised FL methods that require model adaptation.

\subsection{Non-I.I.D settings}
\begin{figure}[h]
    \centering
    \includegraphics[width=0.9\columnwidth]{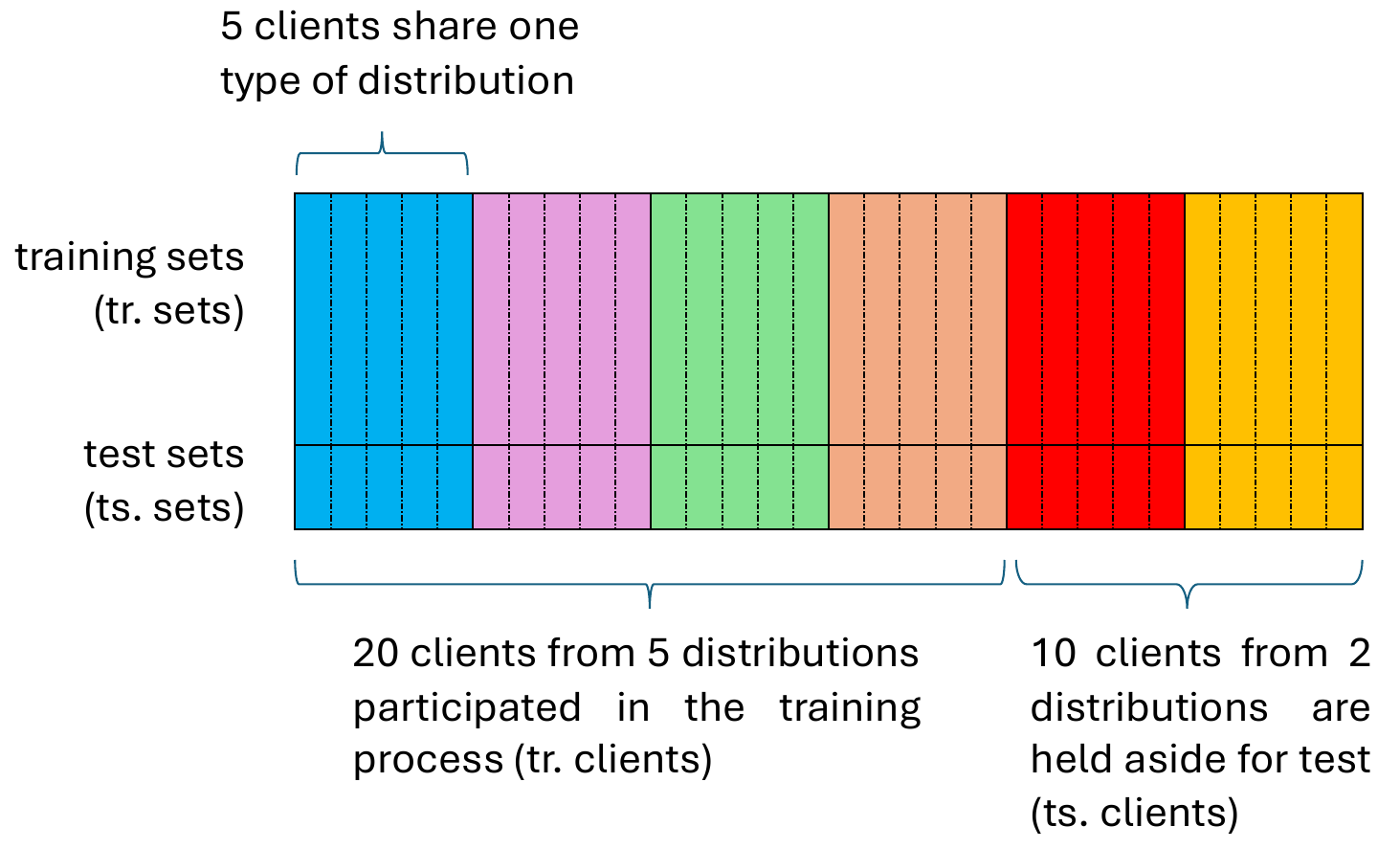}
    \caption{Clients in the target shift setting. Each bar denotes a client. Each colour indicates one type of distribution. Samples on each client are split into a training set and a test set.}
    \label{chpt6:fig4: client settings label shift}
\end{figure}

\begin{figure}[h]
    \centering
    \includegraphics[width=0.9\columnwidth]{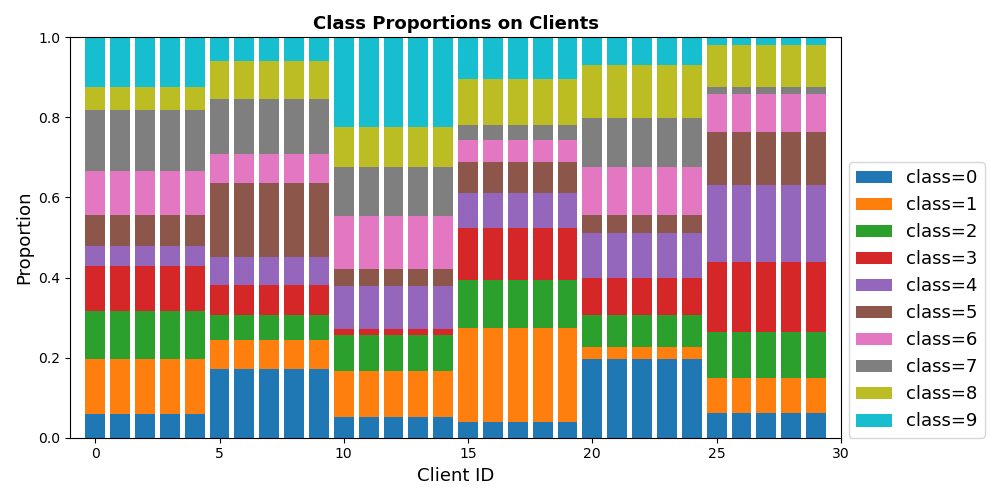}
    \caption{Class distributions on clients. Each bar denotes the class distribution on a client. Each colour corresponds to a class and the length indicates its proportion on the client.}
    \label{chpt6:fig5: class distributions label shift}
\end{figure}

\textbf{Target Shift}: MNIST is applied as a benchmark to simulate the non-I.I.D. environments. The experiment allocates samples of each class individually according to a posterior of the Dirichlet distribution\cite{hsu2019measuring}, which divides clients into five groups with different class distributions. Three groups of clients will participate in the collaborative training process, and the rest will be held for testing. An illustration of client settings is in \textbf{Figure~\ref{chpt6:fig4: client settings label shift}}. Class distributions are shown in \textbf{Figure~\ref{chpt6:fig5: class distributions label shift}}.

\begin{figure}[h]
    \centering
    \includegraphics[width=0.9\columnwidth]{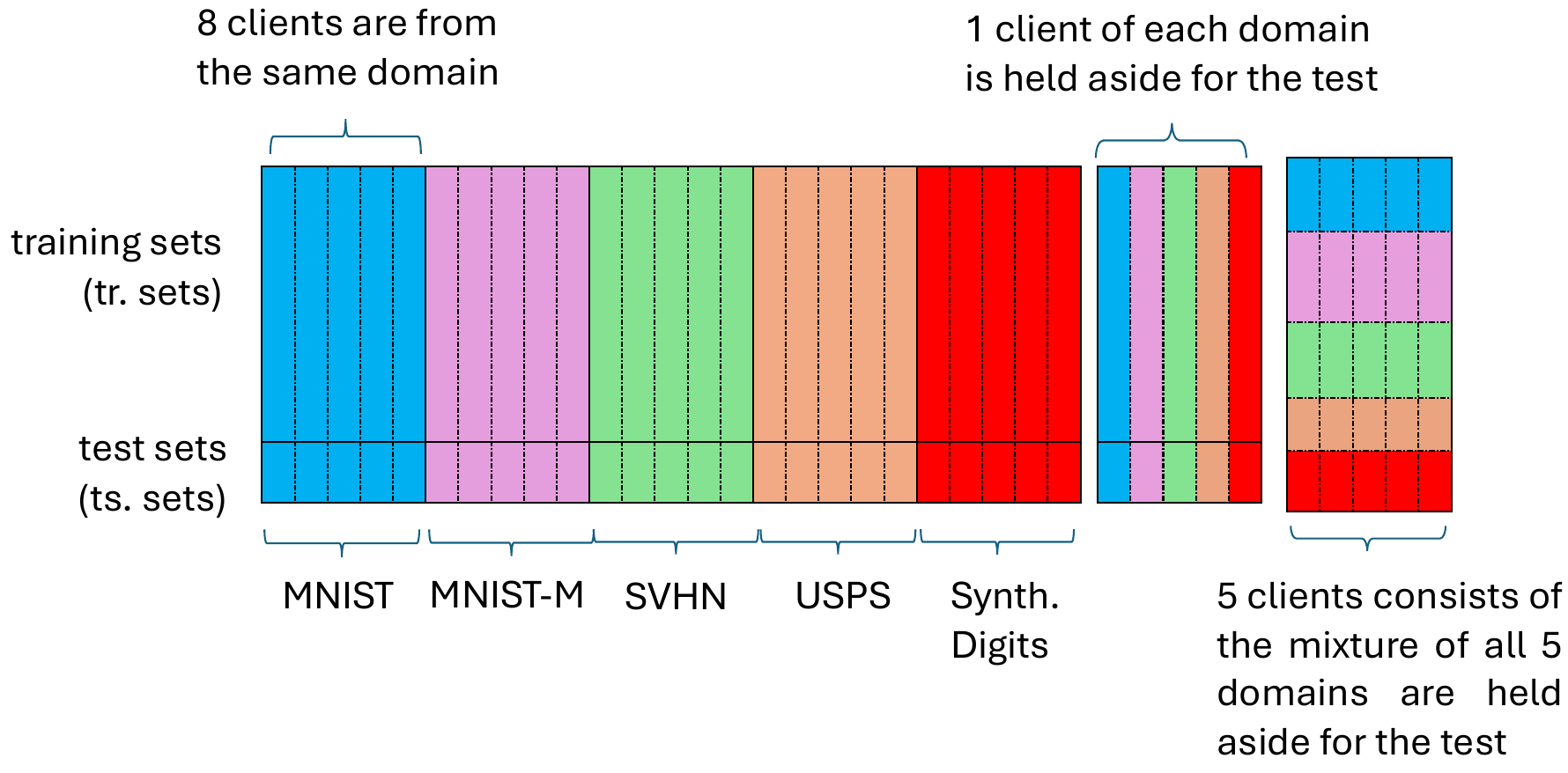}
    \caption{Clients in the feature shift setting. Each bar denotes a client. Each colour indicates a domain. Samples of each client are split into a training set and a test set.}
    \label{chpt6:fig6: client settings feature shift}
\end{figure}

\textbf{Feature Shift}: The research utilises the Digit-5 dataset to evaluate FedVC's performance on feature-shift data. The Digit-5 consists of digits from five different domains (MNIST, MNIST-M, SVHN, USPS and Synth Digits). The experiment assigns samples of each domain to six clients, where five clients will participate in training the global model and one will be held aside for the test.Classes are evenly distributed on each client. In addition, it randomly draws samples from all domains to compose five mixed datasets for the rest clients for the test. An illustration of client settings is in \textbf{Figure~\ref{chpt6:fig6: client settings feature shift}}.

\subsection{Models and Hyperparameters}
The research applies convolution neural networks (CNN) as fundamental models and supervises the training process by virtual concepts. By default, in each communication round, ten clients are sampled to update the global model and virtual concepts, and subsequently, the global model is synchronised to all clients to evaluate its performance. The learning rate of a client's local training step is initialised as 0.005 and it will decay at the rate of 0.8 every 10 communication rounds. During each communication round, a client will tune the global model on its local data for two epochs with a batch size of 10.

For the FedVC, the default number of virtual concepts is set to be 10 and the dimension of each virtual concept is 10. The similarity parameter $\iota$ is 0.1, and the smoothing parameter $\kappa$ is 0.05.

\subsection{Baseline Methods}
Several PerFL strategies are compared as baselines, including: 
\begin{itemize}
    \item \textbf{Local Only}: models those trained on each client locally
    % \item \textbf{FedAvg}: benchmark FL framework ~\cite{mcmahan2017communication}
    \item \textbf{FedAvg + FT}: personalisation by fine-tuning the global model on local data~\cite{cheng2021fine, collins2022fedavg} 
    % \item \textbf{FedAvg + BN}: a global model with shared BatchNormalisation layers
    \item \textbf{FedBN}: a global model with private BatchNormalisation layers~\cite{li2021fedbn}
    \item \textbf{FedProx}: leverages a global to regularise the local training process~\cite{li2018federated}
    \item \textbf{Ditto}: leverages a global to regularise the local training process while learning a local model for each client~\cite{li2021ditto}
    \item \textbf{FedRep}: personalisation by training local classification heads~\cite{pmlr-v139-collins21a}
    \item \textbf{FedDual}: personalisation by training a global and a local feature extractors~\cite{pillutla2022federated}
\end{itemize}

\subsection{Performance}
This section first demonstrates averaged model performance on all clients, which shows that a global model learned with FedVC will achieve comparable performance to other personalised FL methods that require model adaptation. Then, it looks inside the group-wised metrics to evaluate a model's performance on different distributions. Results show that the global model learned with FedVC is more robust to the changing distributions. The learned global model can be directly deployed on test clients without extra adaptations. 

\subsubsection{Target Shift Settings}
For target shift settings, the averaged accuracy, weighted AUC score and weighted F1 score are applied to evaluate model performance\footnote{https://scikit-learn.org/stable/index.html}. \textbf{Table~\ref{cpt6:table:table-averaged-mnist-tr}} and \textbf{Table~\ref{cpt6:table:table-averaged-mnist-ts}} respectively report the averaged performance over the training clients and the test clients. \textbf{Figure~\ref{chpt6:fig7: minist grouped accuracy}} shows the group-wise performance.

\textbf{Overall performance}\\
\textbf{Table~\ref{cpt6:table:table-averaged-mnist-tr}} demonstrates models' performance on the MNIST dataset on the training clients (tr-clients). It can be found that a global model trained by FedVC achieves the best performance under the target shift setting. It outperforms those locally fine-tuned global models (FedAvg+FT) and models with client-specific parameters (FedBN, FedProx, FedRep and FedDual). \textbf{Table~\ref{cpt6:table:table-averaged-mnist-ts}} demonstrates models' performance on the test clients (ts-clients). All baseline methods are fine-tuned on the test clients to adapt to the client's local distribution. It can be found that the model learned by FedVC generalised well to the unseen clients, even though they are not fine-tuned. Note that locally trained models (Local Only and Ditto) can not be generalised to unseen clients.

\begin{table}[t]
\centering
%\resizebox{.95\columnwidth}{!}{
\begin{tabular}{l|c|c|c}
    ~                  & avg. Acc (\%) $\uparrow$ & w. AUC (\%) $\uparrow$     & w. F1 (\%) $\uparrow$ \\
    \hline
    Local Only          & 95.79 (1.00) & 99.69 (0.11) & 93.21 (0.96) \\
    FedAvg+FT           & 97.92 (0.98) & 99.90 (0.04) & 95.49 (1.07) \\
    FedBN               & 98.43 (0.86) & 99.90 (0.04) & 95.71 (0.94) \\
    FedProx             & 98.07 (0.90) & 99.89 (0.04) & 95.48 (0.89) \\
    Ditto               & 95.86 (1.19) & 99.71 (0.09) & 93.25 (1.17) \\
    FedRep              & 93.61 (2.55) & 99.53 (0.18) & 91.05 (2.59) \\ 
    FedDual             & 96.87 (0.99) & 99.84 (0.06) & 94.14 (1.25) \\
    \hline
    FedVC        & \textbf{98.56 (0.56)} & 99.90 (0.05) & 95.83 (0.96) \\ 
    FedVC-sg     & 98.51 (0.62) & \textbf{99.90 (0.03)} & \textbf{95.84 (1.10)}
\end{tabular}
\caption{Overall performance on the MNIST dataset on the training clients. The standard deviation of each metric is reported in parentheses. avg. is the abbreviation of 'averaged' and w. denotes the 'weighted'. The $\uparrow$ denotes that the higher the metric is, the better performance a model achieved, and the best performance is highlighted in bold.}
\label{cpt6:table:table-averaged-mnist-tr}
\end{table}

\begin{table}[t]
\centering
%\resizebox{.95\columnwidth}{!}{
\begin{tabular}{l|c|c|c}
    ~                  & avg. Acc (\%) $\uparrow$ & w. AUC (\%) $\uparrow$     & w. F1 (\%) $\uparrow$ \\
    \hline
    FedAvg+FT           & 98.42 (0.84) & 99.91 (0.04) & 95.71 (0.74) \\
    FedBN               & 98.48 (0.84) & 99.91 (0.04) & 95.64 (0.84) \\
    FedProx             & 98.19 (0.98) & 99.90 (0.04) & 95.53 (0.94) \\
    FedRep              & 88.98 (1.37) & 99.00 (0.24) & 86.11 (1.85) \\ 
    FedDual             & 97.80 (0.51) & 99.88 (0.03) & 95.08 (0.57) \\
    \hline
    FedVC        & \textbf{98.79 (0.62)} & \textbf{99.91 (0.03)} & \textbf{95.97 (0.87)} \\ 
    FedVC-sg     & 98.76 (0.67) & 99.91 (0.04) & 95.92 (0.99)
\end{tabular}
\caption{Overall performance on the MNIST dataset on the test clients. The standard deviation of each metric is reported in parentheses. avg. is the abbreviation of 'averaged' and w. denotes the 'weighted'. The $\uparrow$ denotes that the higher the metric is, the better performance a model achieved, and the best performance is highlighted in bold.}
\label{cpt6:table:table-averaged-mnist-ts}
\end{table}

\textbf{Group-wise performance}\\
\textbf{Figure.\ref{chpt6:fig7: minist grouped accuracy}} shows the averaged accuracy of clients within different groups, i.e., data distributions. It shows that the global model trained by FedVC is more robust among different distributions, and it generalises well to unseen distributions (client groups 4-5). Fluctuation in the learning curves indicates that the fine-tuned models (FedAvg+FT) and models with personalised parameters (FedBN, FedProx, FedDual) are slightly unstable. Locally trained models (Local Only and Ditto) and FedRep have significant performance gaps among clients.

\subsubsection{Feature Shift Settings}
This section demonstrates evaluations in feature shift data. \textbf{Table~\ref{chpt6:table:table-averaged-digit5-tr}} shows that FedVC achieves the best accuracy, AUC and F1 score under this setting. Other models are less robust than FedVC and their performances vary significantly among clients (higher standard deviations). Group-wised performance in \textbf{Figure~\ref{chpt6:fig8: digit5 grouped accuracy}} shows that FedVC has a smaller performance gap between different domains and it is more robust for that there is less fluctuation in the learning curves.
\begin{table}[t]
\centering
%\resizebox{.95\columnwidth}{!}{
\begin{tabular}{l|c|c|c}
    ~           & avg. Acc (\%) $\uparrow$ & w. AUC (\%) $\uparrow$ & w. F1 (\%)$\uparrow$ \\
    \hline
    Local Only         & 74.43 (13.60) & 93.98 (4.53) & 71.34 (13.00) \\
    FedAvg+FT          & 80.92 (9.37)  & 96.75 (2.32) & 77.40 (8.87) \\
    FedBN              & 84.34 (10.61) & 97.49 (2.19) & 80.99 (10.02) \\
    FedProx            & 80.50 (11.32) & 96.46 (2.77) & 76.93 (10.82) \\
    Ditto              & 67.30 (18.50) & 92.10 (6.63) & 63.95 (18.54) \\
    FedRep             & 55.44 (20.40) & 85.55 (11.28) & 51.86 (20.74) \\
    FedDual            & 70.36 (15.59) & 93.20 (5.34) & 66.99 (15.57) \\
    \hline
    FedVC        & 85.42 (8.95) & 97.55 (1.81) & 81.88(8.53) \\
    FedVC-sg     & \textbf{85.82 (8.47)} &\textbf{97.59(1.88)} & \textbf{82.27(7.99)}\\
\end{tabular}
\caption{Overall performance on the Digit-5 dataset on the training clients. The standard deviation of each metric is reported in parentheses. avg. is the abbreviation of 'averaged' and w. denotes the 'weighted'. The $\uparrow$ denotes that the higher the metric is, the better performance a model achieved, and the best performance is highlighted in bold.}
\label{chpt6:table:table-averaged-digit5-tr}
\end{table}

\begin{table}[t]
\centering
%\resizebox{.95\columnwidth}{!}{
\begin{tabular}{l|c|c|c}
    ~           & avg. Acc (\%) $\uparrow$ & w. AUC (\%) $\uparrow$ & w. F1 (\%)$\uparrow$ \\
    \hline
    FedAvg+FT          & 77.85 (7.71) & 96.26 (1.99) & 74.57 (7.42) \\
    FedBN              & 83.30 (7.05) & 97.45 (1.32) & 79.69 (6.67) \\
    FedProx            & 76.90 (7.92) & 96.19 (2.02) & 73.55 (7.31) \\
    FedRep             & 34.85 (18.75) & 73.08 (11.66) & 30.24 (18.39) \\
    FedDual            & 67.15 (11.46) & 92.74 (4.66) & 63.85 (11.62) \\
    \hline
    FedVC        & \textbf{86.20 (5.62)} & 97.61 (1.38) &\textbf{82.92 (5.15)} \\
    FedVC-sg     & 85.10 (5.92) & \textbf{97.68(1.39)} & 81.61 (5.70)
\end{tabular}
\caption{Overall performance on the Digit-5 dataset on the test clients. The standard deviation of each metric is reported in parentheses. avg. is the abbreviation of 'averaged' and w. denotes the 'weighted'. The $\uparrow$ denotes that the higher the metric is, the better performance a model achieved, and the best performance is highlighted in bold.}
\label{chpt6:table:table-averaged-digit5-ts}
\end{table}

\subsection{Ablation Study}
This section evaluates the effectiveness of FedVC through experiments on the Digit-5 dataset. The section first validates virtual concepts' capability as supervised information for personalisation by visualising the distribution of estimated client preferences ($\hat{p}$). Then, it analyses the behaviours of hyperparameters by ablation experiments.

\subsubsection{Interpreting Personalisation}
\begin{figure}[h]
    \centering
    \includegraphics[width=1.0\columnwidth]{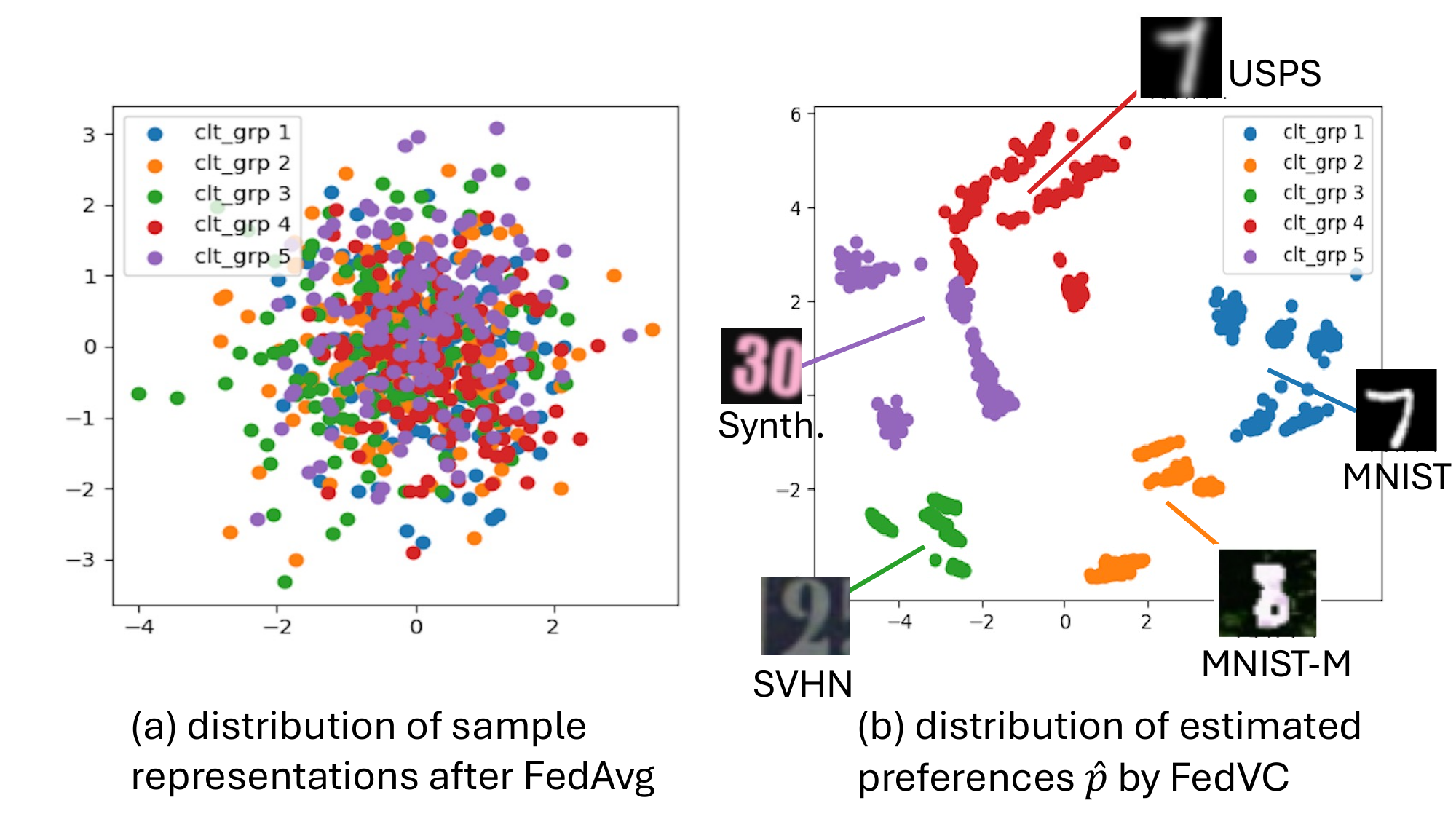}
    \caption{Distribution of estimated client preferences. Colours indicate the client group, i.e., the domain, samples belong to. (a) The aggregation process by vanilla FedAvg will eliminate the information on client preferences so that sample representations are mixed regarding their domains. (b) Virtual concepts succeed in supervising the learning process with client preferences so that the distribution of the estimated client preferences $\hat{p}$ are consistent with their domain knowledge, i.e., samples from the same domain will be closer to each other.}
    \label{chpt6:fig9: distribution of vc}
\end{figure}

\textbf{Figure~\ref{chpt6:fig9: distribution of vc}} compares the latent representations learned by FedAvg and the FedVC. It can be found that FedVC succeeds in supervising the learning process with client preferences so that the distribution of the estimated client preferences $\hat{p}$ are consistent with the group truth knowledge, i.e., samples from the same group (colours) are closer to each other. 

\begin{figure}[h]
    \centering
    \includegraphics[width=1.0\columnwidth]{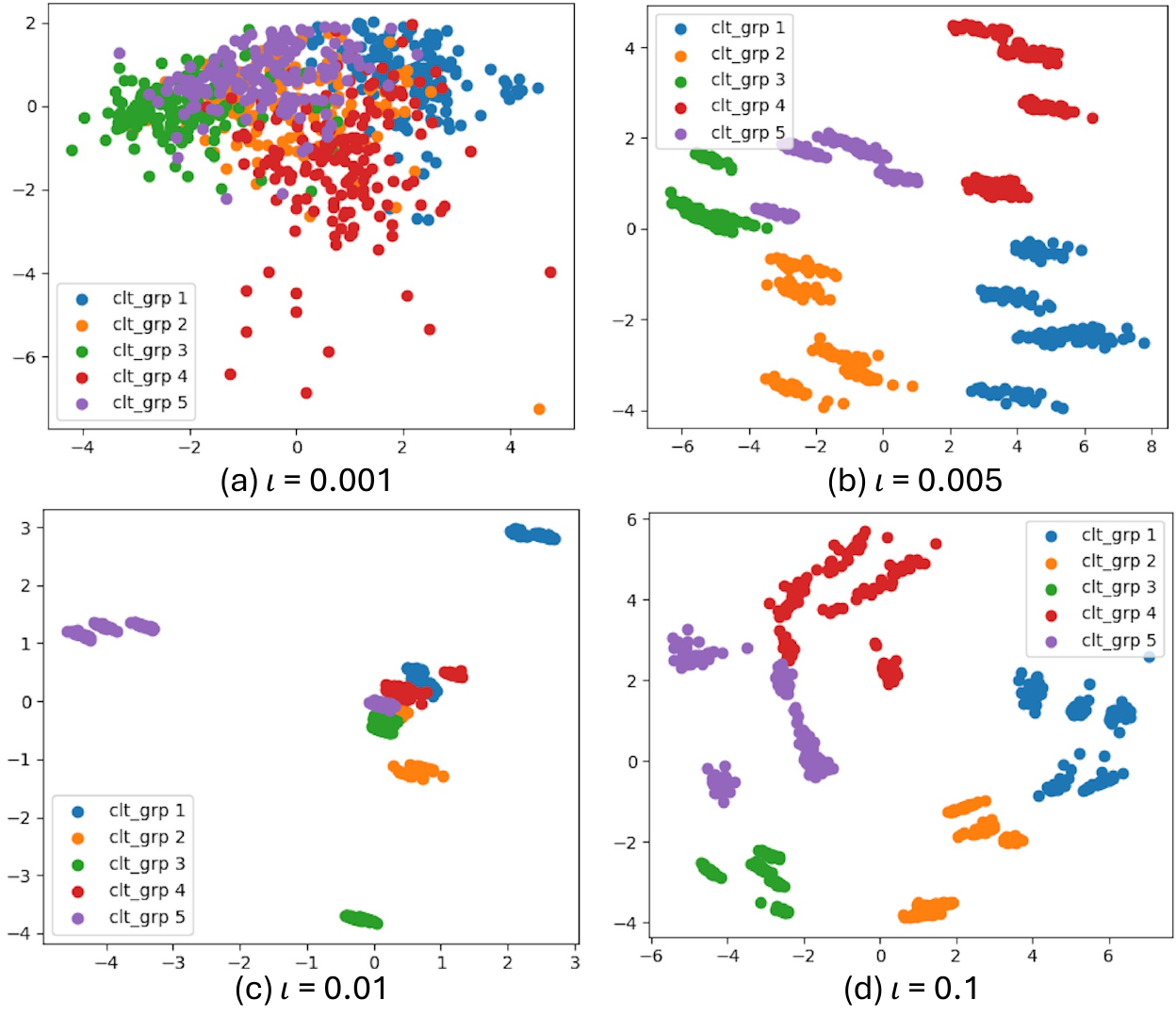}
    \caption{Distribution of estimated client preferences with different $\iota$. The smaller the $\iota$ is, the less weight the difference $|\hat{z}-c|$ when estimating the client preference $\hat{p}$.}
    \label{chpt6:fig10: distribution of vc with iota}
\end{figure}

\begin{figure}[h]
    \centering
    \includegraphics[width=0.9\columnwidth]{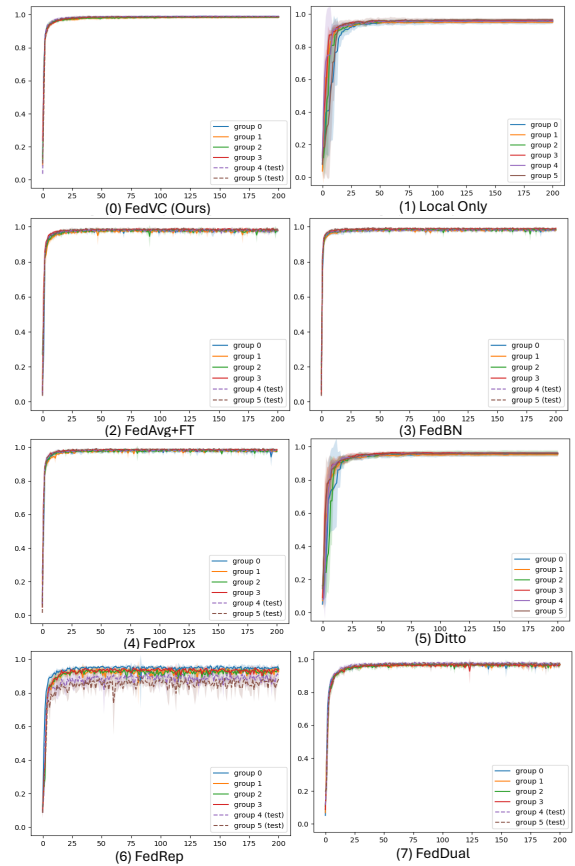}
    \caption{Grouped-wise accuracy on MNIST. The horizontal axis denotes communication rounds and the vertical axis denotes the accuracy. Each colour corresponds to a client group, i.e., data distribution. Shade indicates the standard deviation of accuracy among clients in the group.}
    \label{chpt6:fig7: minist grouped accuracy}
\end{figure}

\begin{figure}[h]
    \centering
    \includegraphics[width=0.9\columnwidth]{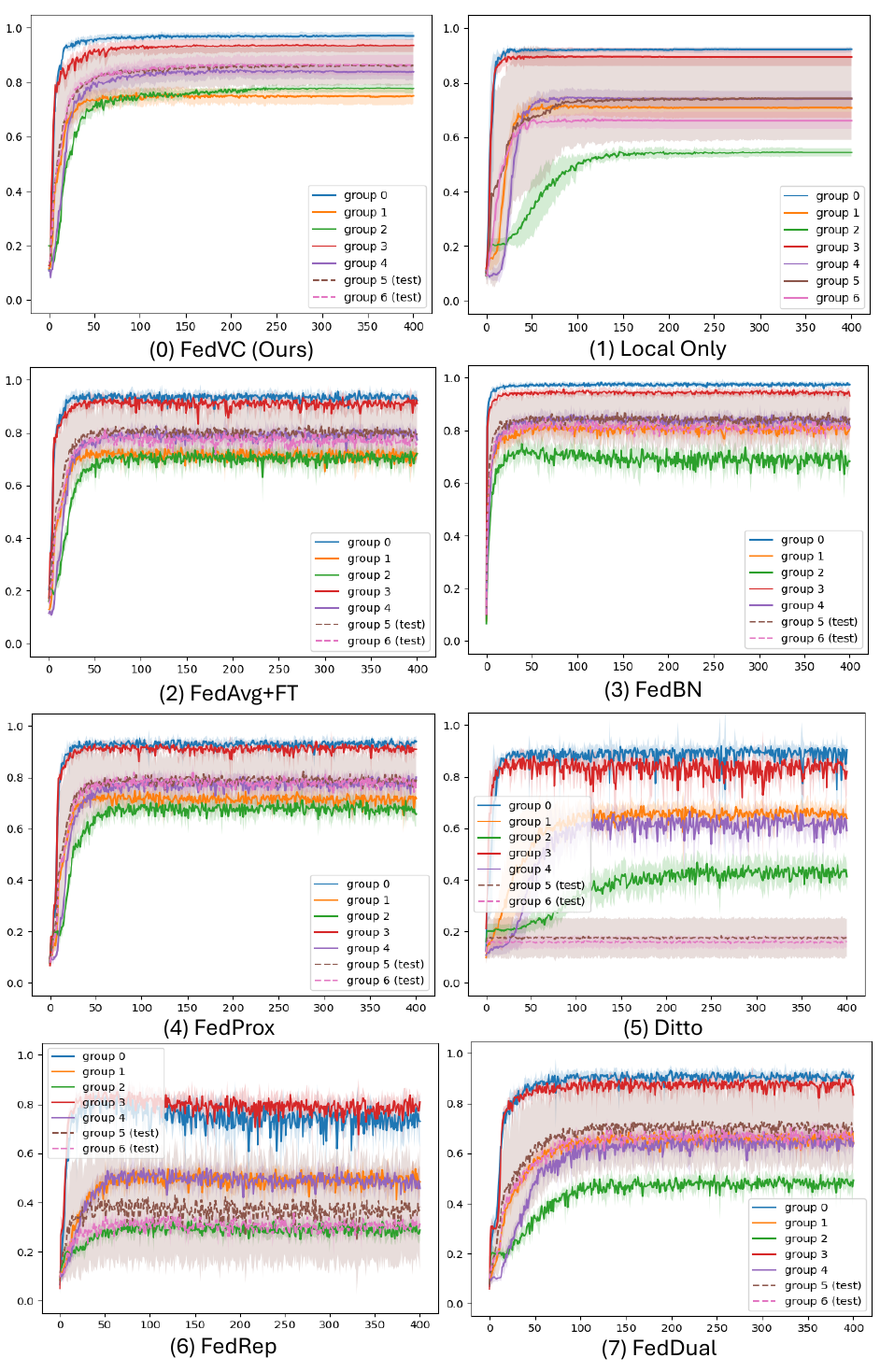}
    \caption{Grouped-wise accuracy on Digit-5. The horizontal axis denotes communication rounds and the vertical axis denotes the accuracy. Each colour corresponds to a client group, i.e., data distribution. Shade indicates the standard deviation of accuracy among clients in the group.}
    \label{chpt6:fig8: digit5 grouped accuracy}
\end{figure}

$\iota$ in \textbf{Equation~\ref{eq: sample relevance to vc}} is a hyperparameter that weights the importance of the difference $|\hat{z}-c|$ when estimating the client preference $\hat{p}$. \textbf{Figure~\ref{chpt6:fig10: distribution of vc with iota}(a)} shows that client preferences (colours) are unrecognisable with a model learned with a small $\iota$, i.e., $\iota=0.001$. With the increasing of $\iota$, the estimated $\hat{p}$ demonstrates structure consistent with their client preferences (\textbf{Figure~\ref{chpt6:fig10: distribution of vc with iota}(b-d)}). It validates the effectiveness of the supervision of virtual concepts $c$. The superior performance of FedVC denotes such supervision does improve the performance of a global model, and virtual concepts are indicators that can be utilised to interpret personalisation.

\subsubsection{Hyperparameters}
The experiments study a hyperparameter's behaviours by evaluating model performance under different values of the selected hyperparameter while holding the others with default values. According to \textbf{Table~\ref{chpt6:table:ablation:n_vc}} and \textbf{Table~\ref{chpt6:table:ablation:d_vc}}, model performance will be improved along with the increasing of the number and the dimension of virtual concepts. \textbf{Table~\ref{chpt6:table:ablation:iota}} shows that a larger weight for the similarity between $\hat{z}$ and $c$ will increase model performance, which validates the effectiveness of the supervision from virtual concepts. In addition, \textbf{Table~\ref{chpt6:table:ablation:kappa}} indicates that the newly estimated $S$, $C$ and $N$ will outperform the older one when using the moving average strategy. \textbf{Table~\ref{chpt6:table:ablation:gamma}} suggests that $\gamma$ needs to be carefully selected when balancing updating the global model and the virtual concepts.

\begin{table}[t]
\centering
%\resizebox{.95\columnwidth}{!}{
\begin{tabular}{c|c|c}
    \# of VCs & avg. Acc (\%)$\uparrow$ on tr & avg. Acc (\%)$\uparrow$ on ts\\
    \hline
    3	& 85.22(9.47) & 85.05(6.79) \\
    6	& 85.24(9.10) & 85.15(6.06) \\
    10  & \textbf{85.42(8.95)} & \textbf{86.20(5.62)}
\end{tabular}
\caption{Performance with different number of virtual concepts}
\label{chpt6:table:ablation:n_vc}
\end{table}

\begin{table}[t]
\centering
%\resizebox{.95\columnwidth}{!}{
\begin{tabular}{c|c|c}
    $d$-VC & avg. Acc (\%)$\uparrow$ on tr & avg. Acc (\%)$\uparrow$ on ts\\
    \hline
    3	& 83.46(9.89) & 83.45(6.95) \\
    6	& 84.36(9.75) & 83.80(7.07)\\
    10  & \textbf{85.42(8.95)} & \textbf{86.20(5.62)}
\end{tabular}
\caption{Performance with different dimensions of virtual concepts}
\label{chpt6:table:ablation:d_vc}
\end{table}

\begin{table}[t]
\centering
%\resizebox{.95\columnwidth}{!}{
\begin{tabular}{c|c|c}
    $\iota$ & avg. Acc (\%)$\uparrow$ on tr & avg. Acc (\%)$\uparrow$ on ts\\
    \hline
    0.001 & 84.40(9.30) & 85.00(6.34) \\
    0.005 & 84.56(9.55) & 85.35(6.44) \\
    0.01 & \textbf{85.74(9.12)} & 85.75(6.25) \\
    0.1	& 85.42(8.95) & \textbf{86.20(5.62)} \\
\end{tabular}
\caption{Performance with different similarity parameter $\iota$. The larger the $\iota$ is, the more weight the difference $|\hat{z} - c|$ when estimating the client preference $\hat{p}$}
\label{chpt6:table:ablation:iota}
\end{table}

\begin{table}[t]
\centering
%\resizebox{.95\columnwidth}{!}{
\begin{tabular}{c|c|c}
    $\kappa$ & avg. Acc (\%)$\uparrow$ on tr & avg. Acc (\%)$\uparrow$ on ts\\
    \hline
    0.01          & \textbf{85.66(8.51)} & 85.40(6.12) \\
    0.05          & 85.42(8.95)	& \textbf{86.20(5.62)} \\
    0.1           & 84.96(9.23) & 85.45(6.18) \\
    0.5           & 84.44(9.16) & 84.65(6.25)\\
    0.95          & 84.02(9.51) & 83.85(7.29)
\end{tabular}
\caption{Performance with different smoothing parameter $\kappa$. The larger the $\kappa$ is, the more weight the previous estimation of $S$, $C$ and $N$.}
\label{chpt6:table:ablation:kappa}
\end{table}

\begin{table}[t]
\centering
%\resizebox{.95\columnwidth}{!}{
\begin{tabular}{c|c|c}
    $\gamma$ & avg. Acc (\%)$\uparrow$ on tr & avg. Acc (\%)$\uparrow$ on ts\\
    \hline
    0.01	& 83.14(10.47)	& 83.40(7.27) \\
    0.1     & 85.24(8.97)	& \textbf{85.30(6.86)} \\
    0.5	    & 83.48(10.63)	& 82.35(8.03) \\
    0.95	& \textbf{85.46(8.71)}	& 85.20(6.02)
\end{tabular}
\caption{Performance with different balancing parameter $\gamma$. The larger the $\gamma$ is, the more important the loss $l_{p}$ to optimising the virtual concepts $c$.}
\label{chpt6:table:ablation:gamma}
\end{table}

\begin{algorithm}
\caption{FedVC}
\label{chapter6:alg1: FedVC}
\textbf{Input:}  communication rounds $R$, epochs in each round $E$, learning rate $\lambda$, batch size $B$, hyperparameters $\iota$, $\kappa$ and $\gamma$\\
\textbf{Output:} optimal parameters $\omega^{*}$, virtual concepts $\mathcal{C}^{*}$
\begin{algorithmic}[1]
\State \textbf{server} initialises parameters $\omega$ and virtual concepts $\mathcal{C}$
\For{$r$ from $0$ to $R$}\Comment{communication rounds}
\State \textbf{server} selects a set of clients $\mathbb{C}$
\For{$k\in\mathbb{C}$ \textbf{parallel}}
\State \textbf{client} $k$ synchronises $\omega$ and $\mathcal{C}$ from the \textbf{server}\Comment{network traffic}
\State $\omega_{k}, S_m^{'(k)}, C_m^{'(k)} \gets$\textbf{ClientUpdate($\omega$)}
\EndFor
\State \textbf{server} collects local updates $\omega_{k}$, $S_m^{'(k)}$ and $C_m^{'(k)}$   $k\in\mathbb{C}$\Comment{network traffic}
\State $\omega\gets\sum_{k\in\mathbb{C}}\alpha_{k}\omega_{k}$
\State update $c_m\in\mathcal{C}$ by \textbf{Equation~\ref{eq: update c_m moving average}}
\EndFor
\State \textbf{return} $\omega$, $\mathcal{C}$
\end{algorithmic}
\textbf{ClientUpdate($\omega$, $\mathcal{C}$)}
\begin{algorithmic}[1]
\For{any sample on the clients} \Comment{Update client preferences$p^{(k)}$}
\State get model outputs by $\hat{y}, \hat{z}=f(x;\omega)$
\State calculate $s_{i,m}^{(k)}$ by \textbf{Equation~\ref{eq: sample relevance to vc}}
\State update $v_m^{(k)}$ by \textbf{Equation~\ref{eq: update upsilon moving average}}
\State update client preference $p^{(k)}\gets\sum_{m=1}^{M}v_{m}^{(k)}c_{m}$
\EndFor
\For{$e$ from $0$ to $E$}
    \For{$b$ from $0$ to $N_k/B$}
        \State sample a batch of data $\mathbb{B}$
        \State $\omega\gets\omega-\nabla_{\omega}(l_p+l_{cls})$ \Comment{Update model}
        \State update $S$, $C$ and $N$ by \textbf{Equation~\ref{eq: S moving average}}, \textbf{\ref{eq: C moving average}} and \textbf{\ref{eq: N moving average}} respectively
    \EndFor
\EndFor
\State \textbf{return} $\omega$, $S$ and $C$ 
\end{algorithmic}
\end{algorithm}

\begin{algorithm}
\caption{FedVC-unified}
\label{chapter6:alg1: FedVC end-to-end}
\textbf{Input:}  communication rounds $R$, epochs in each round $E$, learning rate $\lambda$, batch size $B$, hyperparameters $\iota$, $\kappa$ and $\gamma$\\
\textbf{Output:} optimal parameters $\omega^{*}$, virtual concepts $\mathcal{C}^{*}$
\begin{algorithmic}[1]
\State \textbf{server} initialises parameters $\omega$ and virtual concepts $\mathcal{C}$
\For{$r$ from $0$ to $R$}\Comment{communication rounds}
\State \textbf{server} selects a set of clients $\mathbb{C}$
\For{$k\in\mathbb{C}$ \textbf{parallel}}
\State \textbf{client} $k$ synchronises $\omega$ and $\mathcal{C}$ from the \textbf{server}\Comment{network traffic}
\State $\omega_{k}$, $\mathcal{C}_{k}$ $\gets$ClientUpdate($\omega, \mathcal{C})$
\EndFor
\State \textbf{server} collects local updates$\omega_{k}, \mathcal{C}_{k}$ $k\in\mathbb{C}$\Comment{network traffic}
\State $\omega\gets\sum_{k\in\mathbb{C}}\alpha_{k}\omega_{k}$
\State $c_m\gets\sum_{k\in\mathbb{C}}\alpha_{k}c_m^{(k)}$, $c_m^{(k)}\in\mathcal{C}_{k}$
\EndFor
\State \textbf{return} $\omega$, $\mathcal{C}$
\end{algorithmic}
\textbf{ClientUpdate($\omega$, $\mathcal{C}$)}
\begin{algorithmic}[1]
\For{any sample on the clients} \Comment{Update client preferences$p^{(k)}$}
\State get model outputs by $\hat{y}, \hat{z}=f(x;\omega)$
\State calculate $s_{i,m}^{(k)}$ by \textbf{Equation~\ref{eq: sample relevance to vc}}
\State update $v_m^{(k)}$ by \textbf{Equation~\ref{eq: update upsilon moving average}}
\State update client preference $p^{(k)}\gets\sum_{m=1}^{M}v_{m}^{(k)}c_{m}$
\EndFor
\For{$e$ from $0$ to $E$}
    \For{$b$ from $0$ to $N_k/B$}
        \State sample a batch of data $\mathbb{B}$
        \State $\omega\gets\omega-\nabla_{\omega}\mathcal{L}_k$
        \State $c_m\gets c_m-\nabla_{c}\mathcal{L}_k$, $c_m\in\mathcal{C}$
    \EndFor
\EndFor
\State \textbf{return} $\omega$, $\mathcal{C}$ 
\end{algorithmic}
\end{algorithm}

\section{Conclusions}
The research proposes to utilise virtual concepts as client supervision information to learn a robust global model and to interpret the non-IID data across clients. Specifically, the proposed FedVC interprets each client's preferences as a mixture of conceptual vectors each one represents an interpretable concept to end-users. These conceptual vectors could be learnt via the optimisation procedure of the federated learning system. In addition to the interpretability, the clarity of client-specific personalisation could also be applied to enhance the robustness of the training process on the FL system. The effectiveness of the proposed methods has been validated on benchmark datasets.

\bibliography{aaai24}

\end{document}